\documentclass[10pt, conference, compsocconf]{IEEEtran}
\usepackage{amsmath}
\usepackage{amssymb}
\usepackage{hyperref}
\usepackage{placeins}
\usepackage[numbers,sort&compress,square]{natbib}
\usepackage{multirow}
\usepackage{multicol}
\usepackage{caption}
\usepackage{floatrow}   
\usepackage{subcaption}
\usepackage{graphicx,xcolor} 
\usepackage[framemethod=tikz]{mdframed}
\newcommand{\etal}{\textit{et al}. }

\ifCLASSINFOpdf
\else
\fi
\begin{document}
%
\title{Automatic Temporally Coherent Video Colorization}




%
\author{\IEEEauthorblockN{Harrish Thasarathan,
Kamyar Nazeri,
Mehran Ebrahimi}
\IEEEauthorblockA{Imaging Lab, Faculty of Science\\
University of Ontario Institute of Technology,
Canada\\ harrish.thasarathan@uoit.net, kamyar.nazeri@uoit.ca, mehran.ebrahimi@uoit.ca}}


\maketitle

\begin{abstract}
	Greyscale image colorization for applications in image restoration has seen significant improvements in recent years. Many of these techniques that use learning-based methods struggle to effectively colorize sparse inputs. With the consistent growth of the anime industry, the ability to colorize sparse input such as line art can reduce significant cost and redundant work for production studios by eliminating the in-between frame colorization process. Simply using existing methods yields inconsistent colors between related frames resulting in a flicker effect in the final video. In order to successfully automate key areas of large-scale anime production, the colorization of line arts must be temporally consistent between frames. This paper proposes a method to colorize line art frames in an adversarial setting, to create temporally coherent video of large anime by improving existing image to image translation methods. We show that by adding an extra condition to the generator and discriminator, we can effectively create temporally consistent video sequences from anime line arts. 
\end{abstract}



%
\IEEEpeerreviewmaketitle

\section{Introduction}
In recent years the popularity of anime, Japanese books and cartoons, has risen significantly. The Japanese animation market has recorded major growth for  seven consecutive years crossing 18 billion dollars in revenue with the largest growth occurring in movies and internet distribution in 2017 \cite{Anime:Industry}. The production of anime is a multistaged process that requires time and effort from multiple teams of artists. The drawing of key frames that define major character movements are done by lead artists while in-between frames that fill in these motions are animated by inexperienced artists. The colorization of line art sketches from key frames and in-bewteen frames is considered to be tedious, repetitive, and low pay work. Thus, finding an automatic pipeline for consistently producing thousands of colored frames per episode from line art frames can save significant expenses for anime production studios while simultaneously expediting the animation process. Several approaches exist that address the problem of image colorization. Non learning-based methods often use reference images to match luminance or color histograms \cite{Colorizationnondeep, Colorizationnondeep2} while learning-based methods use Convolutional Neural Networks (CNNs) and treat the problem as a classification task at the pixel level. These techniques take greyscale images as input and do not consider sparse inputs such as anime line art drawings. In addition, when used to form stitched-together video from individual frames, the video lacks consistency between frames. Creating a method to colorize frames to coherent video sequences is crucial for large-scale anime production. Methods that target anime line art colorization rely heavily on learned priors, or color hints to generate a single colored image \cite{Anime:User-guided,Anime:Style-Transfer}. These methods do not scale to colorization of frame sequences and instead target the colorization of generalized single line art drawings to cater to a broad audience of artists and enthusiasts.   

\begin{figure}
    \centering
    \includegraphics[width=\textwidth]{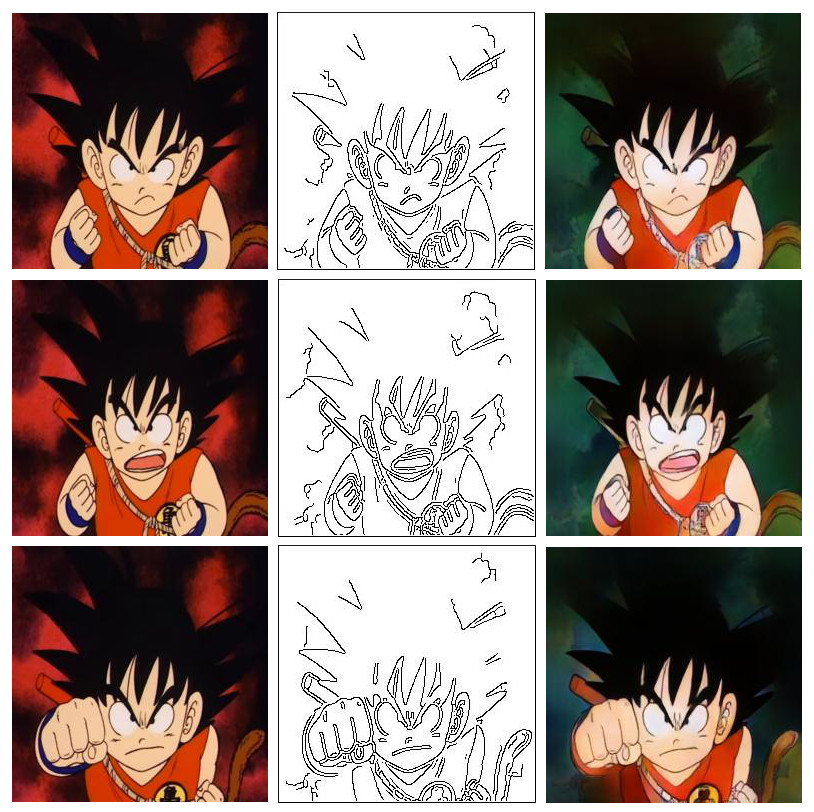}
    \caption{(Left) Ground truth colored image. (Center) Computed synthetic line art. (Right) Generated colorized frame from line art. }
    \label{fig:my_label}
\end{figure}

The image to image translation method presented by Isola \etal \cite{Pix2pix} uses Generative Adversarial Networks (GANs) conditioned on some input to learn a mapping from input to target image by minimizing a loss function. The mapping learned from the proposed technique works with some sparse inputs and has been used in applications such as converting segmentation labels to street view images. We aim to extend this model to work with synthetic line arts. Our proposed model takes temporal information into account to encourage consistency between colorized frames in video form. Our source code is available at: \url{https://github.com/Harry-Thasarathan/TCVC}




\section{Related Works}
Image to image translation using conditional GANs \cite{goodfellow2014generative,odena2018generator} is especially effective in comparison to CNN-based models for colorization tasks \cite{Nazeri:CGAN}. This model successfully maps a high dimensional input to a high dimensional output using a U-Net \cite{U-Net} based generator and patch-based discriminator \cite{Pix2pix}. The closer the input image is to the target, the better the learned mapping is. As a result, this technique is particularly suited to colorization tasks. The U-Net architecture acts as an encoder decoder to produce images conditioned on some input. The limitation with U-Net is the information bottleneck that results from downsampling and then upsampling an input image. Skip connections copy mirrored layers in the encoder to the decoder, but downsampling to 2x2 can lose information. This is especially relevant when considering the sparse nature of line art in comparison to greyscale images. Downsampling input data that is already sparse to that extent should be avoided in the context of anime line art colorization due to the risk of data loss.


The neural algorithm for artistic style presented by Gatys \etal \cite{Style:Gatys} provides a method for the creation of artistic imagery. This is highly relevant as it demonstrates a way to learn representations of both content and style between two images using the pretrained VGG network \cite{simonyan2014very}, then transfer that learned representation with iterative updates on a target image. Johnson \etal \cite{Style:Johnson} showed that this model is optimal for transferring a learned representation of style from a painting which includes encoding texture information to an input photo.  Although this method alone is not effective in transferring color to inputs like line art for our specific task, the ability to learn and differentiate style and content using a pretrained network is highly useful \cite{gondal2018unreasonable}. 

There are very few existing methods for anime line art colorization. Existing methods are made to be highly generalizable for use by different artists \cite{Anime:Paintschainer}. As a result, they rely on user provided color hints to colorize individual line art drawings that are not related. For large-scale anime production, providing color hints for 3000 frames of an episode should be avoided as it will be monotonous work requiring significant user intervention. Additionally, learning a representation for a show based on previous already colored episodes will be more useful for building a method to create temporally coherent frames for future episodes and seasons. 


\section{Method}
Our method attempts to colorize line art frames for large-scale anime production settings by taking into consideration temporal information to account for consistency between frames. We also try to minimize the amount of user intervention required to create colorized frames to mitigate the monotonous workload otherwise required by artists. 

\subsection{Loss Objective}
Let $G$ and $D$ represent the generator and discriminator of our colorization network. Our generator takes a greyscale or line art image $\mathbf{I}_{line}$ as input conditioned on the previous color frame $\mathbf{F}_{prev}$ and returns a color prediction temporally consistent to the previously colored frame 
\begin{equation}
    \mathbf{F}_{pred} = G(\mathbf{I}_{line},\mathbf{F}_{prev}).
\end{equation}
A joint loss is used to train the network that takes advantage of both conditional GAN and neural style algorithm that consists of an adversarial loss, style loss, content loss, and $l_{1}$ loss. The adversarial loss is defined in equation (\ref{eq:g_adv}) and we include $\mathbf{F}_{prev}$ as an additional condition to encourage temporal consistency.
\begin{multline}
		\mathcal{L}_{adv} =\mathbb{E}_{(\mathbf{I}_{line},\mathbf{F}_{prev})} \left[ \log D(\mathbf{I}_{line},\mathbf{F}_{prev}) \right] \\
		+ \mathbb{E}_{\mathbf{F}_{prev}} \log \left[ 1 - D(\mathbf{F}_{pred}, \mathbf{F}_{prev}) \right].
	\label{eq:g_adv}
\end{multline}
We incorporate content and style loss described in \cite{Style:Gatys,Style:Johnson} to further supplement the training of our colorization network. Content loss encourages perceptual similarity while style loss encourages texture similarities between predicted and ground truth color frames. Perceptual similarity is accomplished by minimizing the Manhattan distance between feature maps generated by intermediate layers of VGG-19. Content loss is defined by equation (\ref{eq:l_content}) where $\phi_{i}$ represents the activation map at a given layer $i$ of VGG-19, $\mathbf{F}_{gt}$ is the current ground truth color frame, and $\mathbf{F}_{pred}$ is the generated frame. $N_i$ represents the number of elements in the $i^{th}$ activation layer of VGG-19. For our work, we use activation maps from layers $\tt{relu1\_1, ~relu2\_1, ~relu3\_1, ~relu4\_1}$ and $\tt{relu5\_1}$ 

\begin{equation}
    \mathcal{L}_{content} = \mathbb{E}_i \left[\frac{1}{N_i} \left\lVert \phi_i (\mathbf{F}_{gt}) - \phi_i (\mathbf{F}_{pred}) \right \rVert_1 \right].
\label{eq:l_content}
\end{equation}
Style loss is calculated in a similar manner but rather than calculating the Manhattan distance between feature maps, we calculate the distance between the gram matrices of the feature maps

\begin{equation}
   \mathcal{L}_{style} = \mathbb{E}_j \left[ \lVert G_j^{\phi} ({\mathbf{F}}_{pred}) - G_j^{\phi} ({\mathbf{F}}_{gt}) \rVert_1 \right].
\label{eq:l_style}
\end{equation}

The Gram matrix of feature map $\phi$ is defined by $G^{\phi}_{j}$ in equation (\ref{eq:l_style}) which distributes spatial information containing non localized information such as texture, shape, and style. We also add an $l_{1}$ term to our overall loss objective to preserve structure and encourage the generator to produce results similar to our ground truth. We use adversarial loss with $l_{1}$ to produce sharper generated outputs. The resulting final loss objective is the following 
\begin{equation}
    \mathcal{L} = \lambda_{adv}\mathcal{L}_{adv} + \lambda_{content}\mathcal{L}_{content}+ \lambda_{style}\mathcal{L}_{style}+ \lambda{l_{1}}\mathcal{L}_{l_{1}}.
\end{equation}

 For our experiments $\lambda_{adv}$ = $\lambda_{content}$ = 1, $\lambda_{style}$ = 1000 and $\lambda_{l1}$ = 10. We do not incorporate content loss for greyscale experiments since the content of a greyscale and colored image are the same.

\begin{figure*}
	\centering
	\includegraphics[height=.14\textheight]{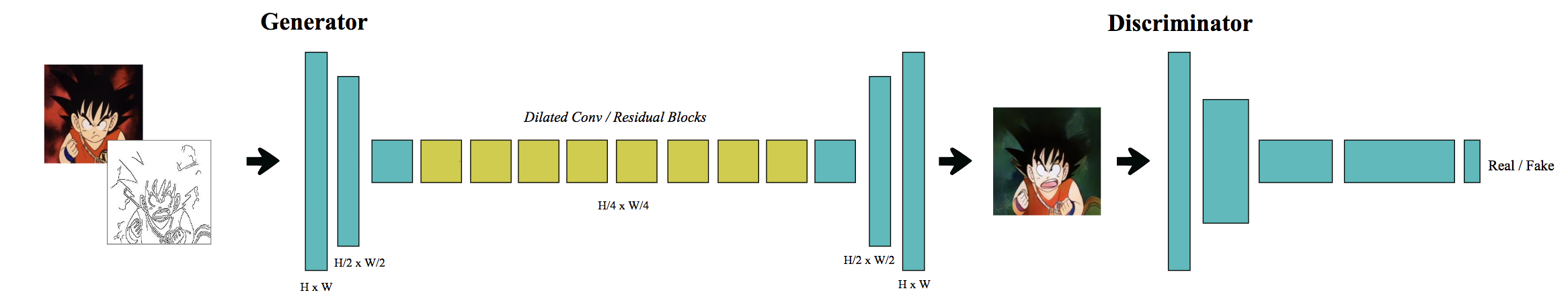}
	\caption{Summary of our proposed method. Lineart frame using Canny edge detection \cite{canny1986computational} and previous color frame are passed as conditions to our generator. Colorized lineart output is classified by the discriminator as real or fake for each iteration during training.}
	\label{seq3}
\end{figure*}

\subsection{Network Architecture}
Our generator is comprised of 2 downsampling layers and 8 residual blocks \cite{residualblocks} that circumvent the bottle-neck issue raised when using a U-Net-based architecture. U-Net downsamples the input image to 2x2 which adversely affects already sparse input data. On the other hand, residual blocks reduce the need to downsample, instead allowing layers to be skipped in the training process. A simpler residual function is learned versus a function that directly maps input line art to colored frames, where each frame is a single image. Instance normalization \cite{Norm:Instance} is used in place of batch normalization since smaller batch sizes are used for memory requirements when conditioning two images. With smaller batches, it is more effective to normalize across the spatial dimension alone versus the spatial and batch dimension for all images. The output of the residual blocks is then upsampled to the original input size. Our discriminator architecture uses a 70x70 patch GAN as in Isola \etal \cite{Pix2pix} who showed mapping generator predictions to a scale of $N\times{N}$ outputs and classifying each patch as real or fake can be more effective than regular discriminators that map to a single scalar. Our discriminator also takes advantage of spectral normalization to stabilize training as shown in \cite{Norm:Spectral}. By restricting the Lipschitz constant of the discriminator to 1, we prevent the discriminator from learning a representation that perfectly differentiates real from fake generated images. Instead the discriminator is encouraged to find the intended optimal solution.     

In Chan \etal \cite{Dance:Effros} temporal information is introduced in a cGAN architecture by adding extra conditions on their generator and discriminator. The current input and previously generated image are conditions for their generator. The discriminator is then conditioned on the current and previous input, generated image, and ground truth. We take inspiration from this method and incorporate temporal information in our network with added conditions to our generator and discriminator. The conditions used by Chan \etal \cite{Dance:Effros} can be computationally expensive for our resources so we choose to use the line art frame and the previous ground truth color frame as conditions to our generator. The first frame of the training set does not have a previous frame, so we pass a blank image as the condition to the network. In every other case we randomly sample either the blank image or the corresponding previous color frame according to a Bernoulli distribution. Thus there is a 50 percent chance that any given condition $\mathbf{F}_{prev}$ is a blank image. 

\subsection{Experiments}
\subsubsection{Implementation Strategy}
In order to show viability for large-scale productions, we choose to make our dataset from the anime Dragonball which has nearly 300 episodes spread over 9 seasons. Our dataset are 2 full seasons obtained from legal sources. In order to extract frames, we use OpenCV to save frame sequences. We thereby create a training set of 84,000 frames and a test and validation set of just over 15,000 frames that make up multiple episodes previously unseen by our network at inference. Our input pipeline converts RGB frames to both greyscale and line art for separate experiments. We convert the colored ground truth to a single channel greyscale input and mimic line art using Canny edge detection \cite{canny1986computational} with gaussian filter of standard deviation 1. 

Our proposed method is implemented in PyTorch with input frames resized to $256\times{256}$. For every experiment we feed our network batch sizes of 16 for 35 epochs. The generator is trained with a learning rate of $10^{-4}$ while the discriminator trains with a one tenth learning rate of the generator using the Adam optimizer algorithm \cite{Adam}. By doing so, the generator is given the opportunity to learn a mapping before the discriminator becomes too strong which prevents any useful training. Our baseline for experiments was a pix2pix U-Net-based conditional GAN that was only conditioned on line art without considering temporal information, style, and content loss. For both our method and baseline we conduct two separate experiments, one on colorizing greyscale frames and the other on colorizing line art frames. Greyscale images are the most often used medium for colorization related tasks so we include them in our experiments. Being able to colorize greyscale images effectively can also be useful for Japanese comics known as manga. Manga are traditionally greyscale because it is too expensive to pay artists of a popular series to color thousands of images per book that run more than 300 issues. Although manga take the form of books, being able to account for temporal information is still highly relevant. Every page of a manga can consist of small frames that are correlated with each other. Characters are often seen in multiple frames repeated on a page as they move and interact. Thus maintaining consistency even between images that do not make up a video is necessary for automatic colorization.   

\subsubsection{Validation Metrics}
In order to validate results from our experiments, we employ three different image quality metrics, namely Fr\'echet inception distance (FID) \cite{zhang2018unreasonable}, structural similarity index (SSIM) \cite{wang2004image}, and peak signal to noise ratio (PSNR). Evaluation metrics that assume pixel wise independence like PSNR and SSIM can assign favorable scores to perceptually inaccurate results or penalize scores for visibly large differences that do not necessarily imply low perceptual quality \cite{zhang2018unreasonable}. Especially for the case of anime colorization, differences that do not affect human perception are still favorable results. For this reason, we include FID which has been shown to correlate closely with human perception \cite{heusel2017gans}. FID measures the distance between activations of Inception-v3 \cite{inception} trained on ImageNet \cite{russakovsky2015imagenet} and accounts for both the diversity and visual fidelity of generated samples.

\section{Results}
\subsection{Qualitative Comparison}
\begin{figure}[h]
\centering
\begin{subfigure}{.3\linewidth}
    \centering
    \includegraphics[width=\textwidth]{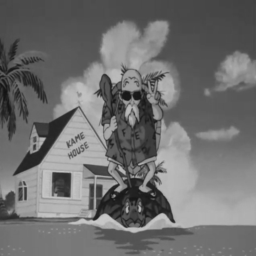}
    \caption{Greyscale}\label{fig:image1}
\end{subfigure}
\begin{subfigure}{.3\linewidth}
    \centering
    \includegraphics[width=\textwidth]{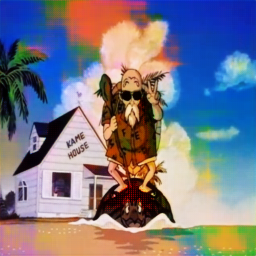}
    \caption{Baseline}\label{fig:image2}
\end{subfigure}
\begin{subfigure}{.3\linewidth}
    \centering
    \includegraphics[width=\textwidth]{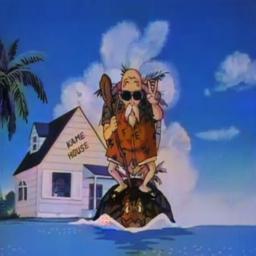}
    \caption{Ours}\label{fig:image3}
\end{subfigure}
\RawCaption{\caption{Comparison of qualitative results from greyscale input. a) Greyscale frame. (b) Baseline (U-Net). (c) Ours.}
\label{GSImg}}
\end{figure}

\begin{figure}[h]
\begin{subfigure}{.3\linewidth}
  \centering
  \includegraphics[width=\textwidth]{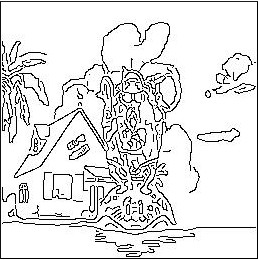}
  \caption{Line art}\label{fig:image4}
\end{subfigure} 
\begin{subfigure}{.3\linewidth}
  \centering
  \includegraphics[width=\textwidth]{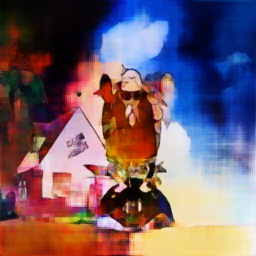}
  \caption{Baseline}\label{fig:image5}
\end{subfigure} 
\begin{subfigure}{.3\linewidth}
  \centering
  \includegraphics[width=\textwidth]{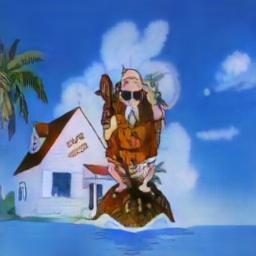}
  \caption{Ours}\label{fig:image}
\end{subfigure} 
\RawCaption{\caption{Comparison of qualitative results from synthesized line art frame input. (a) Synthesized line art frame. (b) Baseline (U-Net). (c) Ours.}
\label{LAImg}}
\end{figure}

\begin{figure*}
	\centering
	\includegraphics[height=.21\textheight]{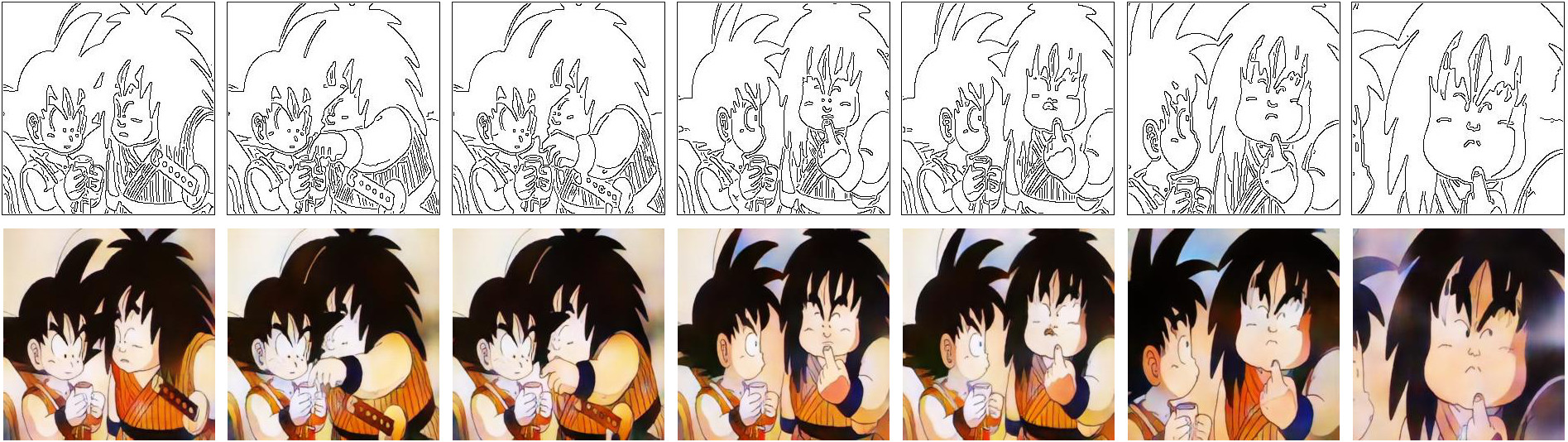}
	\caption{Sequence of colorized frames using our method (bottom) from synthesized 
     line art frames (top). The first frame is conditioned on a blank image with each successive frame being conditioned on the previously generated frame. Any incorrect colorization on the first frame is also present in successive frames.}
	\label{seq1}
\end{figure*}

\begin{figure}
	\centering
	\includegraphics[height=.8\textheight]{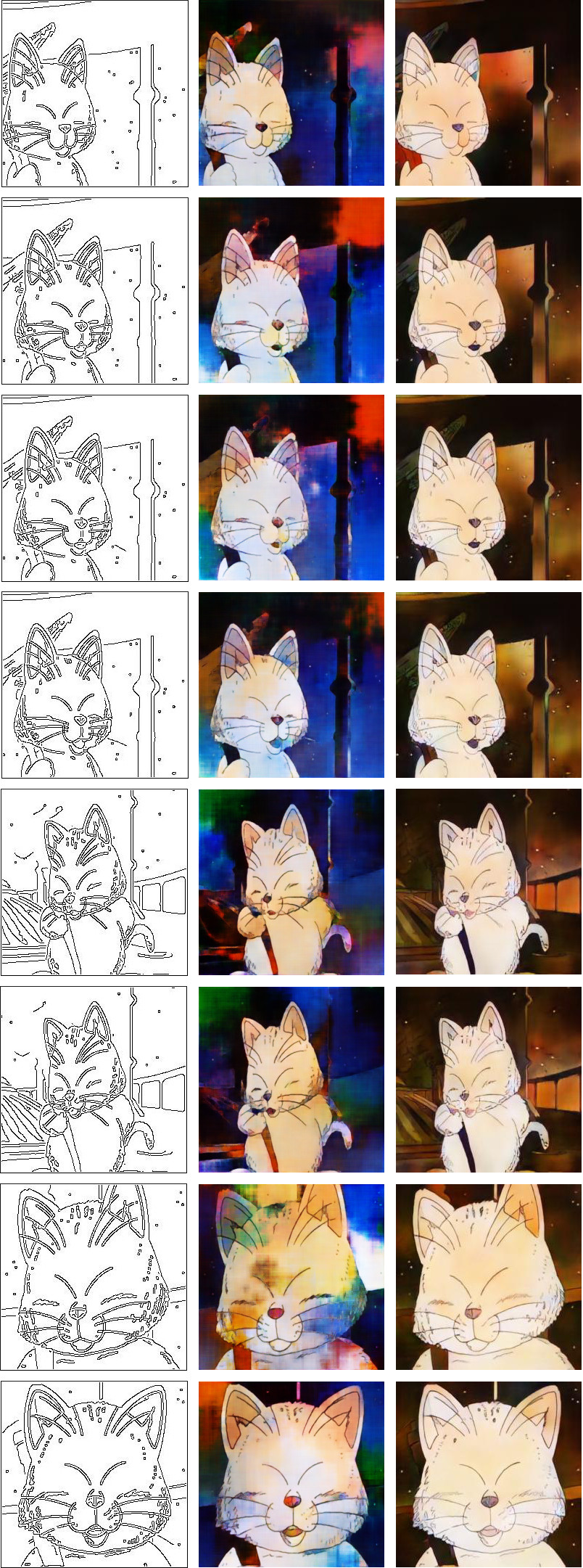}
	\caption{Colorized sequence of frames from line art frames. The first frame is conditioned on a blank image with each successive frame being conditioned on the previously generated frame. (Left) Synthesized line art. (Center) Baseline results. (Right) Our results.}
	\label{seq2}
\end{figure}

Results obtained from our baseline model using U-Net suffered from the checkerboard effect \cite{odena2016deconvolution} especially with line art input as presented in figure \ref{LAImg} likely due to its sparse nature. The greyscale colorization was perceptually natural apart from the checkerboard on some images, however when combining multiple consecutive frames to form a video, the flicker effect from slight variations in colors between frames was very apparent. Inconsistencies between frames resulting in color variations is even more apparent when using line art input in our baseline. Figure \ref{seq2} shows these color inconsistencies that lead to flickering when the frames are stitched together to form a video.  With our model, the checkerboard effect was non existent for line art generated frames even when compared to using greyscale input with our baseline in figure \ref{GSImg}. The content and style loss contributed to learning texture information while reducing unwanted textures such as the checkerboard effect. Additionally, the generated frames of our model from both line art and greyscale created a more coherent video. This is shown both in figure \ref{seq1} and \ref{seq2} where our model produces consecutive frames with less variation which in return reduces flickering in the final video. The flicker effect still exists but is not as significant as our baseline model. 

\subsection{Quantitative Comparison}
Quantitative comparisons are reflected in table \ref{tab1} using PSNR, SSIM, and FID of both models over the validation set. PSNR is higher in our model while the structural similarity index shows our model is closer to the ground truth colored images than our baseline. In terms of human perceptual clarity, the Fr\'echet inception distance between our model that leverages residual blocks with style and content loss is more effective in producing realistic results than the baseline model. 

\begin{table}[h]
\centering
\begin{tabular}{r|c|c|c|c}
\hline
\multirow{3}{*}{Statistic} & \multicolumn{4}{c}{Model}                               \\ 
                           & \multicolumn{2}{c}{Baseline} & \multicolumn{2}{c}{Ours} \\ \cline{2-5}
                           & Greyscale      & Line art     & Greyscale    & Line art   \\ \hline\hline 
FID                        & 20.69          &  32.46            &9.29        & 19.12         \\ \hline
SSIM                       & 0.72            & 0.36       & 0.78    & 0.57 \\ \hline
PSNR                       & 14.15              & 8.74           & 17.38  & 16.03\\ \hline
\end{tabular}
\caption{Quantitative results SSIM, PSNR, and FID between baseline and our model}\label{tab1}
\end{table}

\section{Discussion and Future Work}
Using learning-based methods to map sparse inputs to colored outputs is challenging, however it is evident from our results that it is possible specifically in the domain of anime production. If artists make detailed sketches, they can circumvent the colorization process of key frames and in-between frames thereby expediting the production process while reducing the overall cost. Rather than require color artists to perform the task, the coloring workflow can be adapted such that the number of artists can be reduced, requiring only a few artists to fix regions colored incorrectly. This is true for both anime and manga production. For manga, which is already greyscale, even less effort is needed to automatically colorize frames in each page. Greyscale frames produce results requiring less color correction from artists since more information is present in the input compared to line art. For minimal extra cost, studios can publish colored versions of their most popular manga. 

The main issue that still needs to be addressed is what to do when new characters are introduced. In our case, the validation set consisted of the last four episodes of season 2 in which a new villain is introduced to set up the next season. With our existing model, when episodes that introduce new characters are added, the incorrectly colored characters can be manually colored by artists. When the final episode is finished, they can be added back to the dataset to be trained on the network again. By consistently retraining the model on previously colorized frames adjusted by artists for new characters, future episodes can be correctly colored. Thus, the model can be incorporated into a workflow that circumvents the issue. Canny edge detection sometimes misses the dots used for eyes or the lines for mouths in scenes where characters are distant in the frame. Finding methods to more effectively simulate line art will greatly improve the network in coloring line art frames. 

\section*{Acknowledgment}
This research was supported in part by an NSERC Discovery Grant. The authors gratefully acknowledge the support of NVIDIA Corporation for donation of GPUs through its Academic Grant Program.



%

\end{document}